\documentclass{article}
\usepackage{spconf,amsmath,graphicx,hyperref}
\usepackage{amssymb,color}  
\usepackage{bm}
\usepackage{multirow}  
\usepackage{booktabs}
\usepackage{verbatim}
\usepackage[
  style=ieee,
  backend=bibtex,
  giveninits=true,
  maxbibnames=1,
  minbibnames=1,
  url=false,
  doi=false,
  isbn=false,
]{biblatex}
\newcommand{\bbtheta}{\bm{\theta}}
\newcommand{\bbSig}{\bm{\Sigma}}
\newcommand{\bbphi}{\boldsymbol{\phi}} 
\newcommand{\bbPhi}{\boldsymbol{\Phi}}

\title{Conformalized Gaussian processes for online uncertainty quantification over graphs}
\twoauthors
 {Jinwen Xu, Qin Lu}
	{School of ECE, 
	U. of Georgia}
 {Georgios B. Giannakis}
	{Dept. of ECE,
U. of Minnesota}
\begin{document}
%
\maketitle
\begin{abstract}
Uncertainty quantification (UQ) over graphs arises in a number of safety-critical applications in network science. The Gaussian process (GP), as a classical Bayesian framework for UQ, has been developed to handle graph-structured data by devising topology-aware kernel functions. However, such GP-based approaches are limited not only by the prohibitive computational complexity, but also the strict modeling assumptions that might yield poor coverage, especially with labels arriving on the fly. To effect scalability, we devise a novel graph-aware parametric GP model by leveraging the random feature (RF)-based kernel approximation, which is amenable to efficient recursive Bayesian model updates. To further allow for adaptivity, an ensemble of graph-aware RF-based scalable GPs have been leveraged, with per-GP weight adapted to data arriving incrementally. To ensure valid coverage with robustness to model mis-specification, we wed the GP-based set predictors with the online conformal prediction framework, which post-processes the prediction sets using adaptive thresholds. Experimental results the proposed method yields improved coverage and efficient prediction sets over existing baselines by adaptively ensembling the GP models and setting the key threshold parameters in CP.

\end{abstract}
\section{Introduction}
\label{sec:intro}

A plethora of safety-critical applications in network science entail not only a point prediction for inference in graphs, but also an uncertainty-aware prediction set that can self-assess the quality of the sought prediction. To allow for such uncertainty quantification (UQ) over graphs, Gaussian processes (GPs), the classical Bayesian nonparametric framework, has been extended to handle graph-structured data by defining kernel functions that account for graph topology~\cite{ng2018bayesian, walker2019graph}. These models have demonstrated remarkable performance in applications such as environmental monitoring \cite{krause2008near}, spatial disease prediction \cite{flaxman2015fast}, and real estate valuation \cite{law2019variational}. Recent advances have addressed the computational bottleneck of exact graph-based GPs through various approximation strategies: spectral methods leverage graph Fourier transforms \cite{borovitskiy2021matern, polyzos2021ensemble}, inducing point approaches reduce complexity via sparse representations \cite{wan2023bayesian}, and graph random features achieve $\mathcal{O}(N^{3/2})$ complexity through sparse random walks \cite{zhang2025graphrandomfeaturesscalable}, though the latter operate solely on graph topology without incorporating node features. 

Despite their theoretical elegance,  such graph-adaptive GP methods could produce mis-calibrated uncertainty estimates, if the underlying assumptions (e.g., smoothness, stationarity) are mis-specified, particularly in online learning settings where models continuously adapt to streaming data \cite{ovadia2019can}. While alternative Bayesian graph neural networks based approaches can accommodate non-stationarity and enrich the function expressiveness, they typically require nontrivial approximate Bayesian inference techniques, including Monte Carlo dropout \cite{gal2016dropout}, Bayesian GNNs \cite{zhang2019bayesian}, and deep ensembles \cite{lakshminarayanan2017simple}. Still, they are susceptible to the issue of model mismatch with data arriving in real time. 

To combat against such model-misspecification, conformal prediction (CP) provides a principled {\it post-processing} solution through its distribution-free framework with guaranteed coverage \cite{vovk2005algorithmic, romano2019conformalized}, requiring only the assumption of data exchangeability. Recent graph-based CP methods like DAPS \cite{zargarbashi2023daps} and SNAPS \cite{song2024similarity} have demonstrated CP's effectiveness on graph-structured data by leveraging neighborhood information and similarity-based diffusion. However, these methods rely on rather strict assumption regarding data exchangeability and  use fixed thresholds calibrated on static datasets, limiting their applicability in dynamic environments with distribution shifts. The integration of CP with Graph GPs remains underexplored, particularly adaptive CP variants that can handle online settings where traditional static calibration fails \cite{liu2020gaussian}.

\vspace*{0.1cm}

\noindent{\bf Contributions.} We put forth a framework for UQ on streaming graph data that unifies Graph GP ensembles with online conformal prediction. Our approach leverages Random Fourier Features to achieve linear-time kernel approximation and employs incremental Bayesian updates to handle streaming data efficiently. By pre-computing graph transformations, we incorporate neighborhood structure while preserving real-time update capabilities. The framework combines multiple kernels (RBF variants, Matérn) through ensemble learning and explores various conformal prediction strategies—from traditional fixed-threshold to online adaptive and Bayesian methods. Experiments across synthetic and real-world datasets reveal that kernel ensembling with online threshold adaptation consistently outperforms single-kernel and fixed-threshold baselines in coverage stability, demonstrating that streaming graph data can support both efficient uncertainty quantification and robust statistical guarantees even under distribution shifts.
\section{PRELIMINARIES}
\label{sec:format}
\subsection{GPs over graph-structured data}\label{sec:graph-gp}

Inference in graph-structured data permeates in a number of applications in network science. In this context, consider a graph ${\cal G}:=\{{\cal V}, {\bf A}_N, {\bf X}_N \}$ with $N$ nodes, where the vertex set ${\cal V}:=\{1,\ldots, N\}$ collects all the nodes, ${\bf A}_N\in \mathbb{R}^{N\times N}$ is the adjacency matrix whose $(n,n')$th entry, $a_{n,n'}:={\bf A}_{n,n'}$, denotes the link connecting nodes $n$ and $n'$, and  ${\bf X}_N:=[{\bf x}_1, \ldots {\bf x}_N]$ (${\bf x}_n$ is the $d$-dimensional feature vector for node $n$) is the feature matrix for all the nodes. In addition, each node is associated with a real-valued label $y_n$. Given ${\cal G}$ and the labels over a subset of nodes ${\bf y}_n := [y_1,\ldots, y_n]^\top$, the goal is to infer the label on the unobserved nodes. Here, such a semi-supervised learning (SSL) task will be carried out in an incremental setting, where past observations ${\bf y}_n$ are used to form the predictor of $y_{n+1}$ for node $n+1$, before the new datum $y_{n+1}$ becomes available. In many safety-critical domains, we are not only interested in a {\it point} prediction, but also a {\it set predictor} ${\cal C}_{n+1}$ that can self-assesses the reliability of the sought prediction.

To yield such uncertainty-aware prediction sets, GPs are well-established framework that learns a {\it probabilistic} function mapping $f$ that connects any input $x$ to output $y$. For graph-structured data, efforts have been spent on devising kernel functions that account for the graph topology~\cite{zhi2023gaussian}, of which our focus is on~\cite{ng2018bayesian} given the SSL task. Specifically, a GP prior is postulated for $f({\bf x})$ as: $f\sim {\cal GP}(0, \kappa({\bf x}, {\bf x}'))$ with $\kappa$ being the positive-definite {\it kernel} function that measures pairwise similarity. This implies that,
for all the node features on graphs $\mathbf{X}_{N}$, the joint prior pdf for the function evaluations ${\bf f}_N:=[f({\bf x}_1),\ldots, f({\bf x}_N)]^\top$ will be Gaussian distributed as:  $\mathbf{f}_N \sim \mathcal{N}(\mathbf{0}, \mathbf{K}_N)$, where $[\mathbf{K}_N]_{ij} = \kappa(\mathbf{x}_i, \mathbf{x}_j)$. 

To incorporate the graph relational information, we will rely on the transformed latent variables $\mathbf{h}_N = \mathbf{P}_N\mathbf{f}_N$ where $\mathbf{P}_N = (\mathbf{D}_N + \mathbf{I}_N)^{-1}(\mathbf{I}_N + \mathbf{A}_N)$, with degree matrix $\mathbf{D}_N = \text{diag}({\bf A}_N {\bf 1}_N)$ (${\bf 1}_N$ is an $N\times 1$ all-one vector). The prior pdf of $\mathbf{h}_N$ is then given by \vspace{-0.3cm}
\begin{align}
\mathbf{h}_N | \mathbf{X}_N, \mathbf{A}_N \sim \mathcal{N}(\mathbf{0}, \tilde{\bf K}_N), \quad \tilde{\bf K}_N:= \mathbf{P}_N\mathbf{K}_N\mathbf{P}_N^\top \label{eq:h_N}
\end{align}
For real-valued $y_i$ per node $i$, the connection with the latent variable is characterized by the Gaussian likelihood $p(y_i| h(\mathbf{x}_i)) = \mathcal{N}(y_i; h_i, \sigma_\epsilon^2)$ ($\sigma_\epsilon^2$ is the noise variance). And the batch likelihood is assumed conditionally independent across nodes.  Given observed labels ${\bf y}_n$,  the predictive pdf for $y_{n+1}$ at any node $n+1$ is given by
\begin{align}
p(y_{n+1}|{\cal G},\mathbf{y}_n) = \mathcal{N}(y_{n+1}; \hat{y}_{n+1}, \sigma^2_{n+1}) \label{eq:y_pre}
 \end{align}
where
\begin{subequations}
\begin{align}	
\hat{y}_{n+1} & = \tilde{\mathbf{k}}_{n+1} (\tilde{\mathbf{K}}_n + \sigma_\epsilon^2 \mathbf{I}_n)^{-1} \mathbf{y}_{n}\nonumber \\
\sigma^2_{n+1}& = \tilde{\kappa}_{n+1,n+1} - \tilde{\mathbf{k}}^{\top}({\bf x}) (\tilde{\mathbf{K}}_n + \sigma_\epsilon^2 \mathbf{I}_n)^{-1} \tilde{\mathbf{k}}({\bf x}) + \sigma_\epsilon^2 \nonumber
\end{align}
\end{subequations}
with $\tilde{\mathbf{K}}_n$ being the upper-left $n \times n$ submatrix of the graph-enhanced covariance matrix $\tilde{\mathbf{K}}_N$, $\tilde{\kappa}_{n+1,n+1}:=[\tilde{\mathbf{K}}_N]_{n+1, n+1}$, and $\tilde{\mathbf{k}}_{n+1}:=[[\tilde{\mathbf{K}}_N]_{1,n+1}, \ldots, [\tilde{\mathbf{K}}_N]_{n,n+1}]^\top $. Based on~\eqref{eq:y_pre}, the Bayes $\beta$-credible set is 
\begin{align}
  {\cal K}_{n+1}^\beta = [\hat{y}_{n+1} - z_\beta\sigma_{n+1}, \hat{y}_{n+1} + z_\beta\sigma_{n+1}]   \label{eq:BCS}
\end{align}
where $z_\beta$ is the appropriate quantile based on $\beta$ (e.g., $z_\beta$ = 2 for $\beta = 95\%$). However, the coverage consistency of $ {\cal K}_{n+1}^\beta $ depends on model specification. To achieve robust coverage guarantees under potential model mis-specification, we will integrate this graph-aware GP with the CP framework.

\subsection{CP and adaptation to graphs}
CP is a distribution-free framework~\cite{chen2018discretized} for UQ that is compatible with any prediction model~\cite{vovk2005algorithmic,angelopoulos2023conformal}. Given a prediction model $p(y|{\cal D}_n, {\bf x})$ trained on labeled data ${\cal D}_n:=\{({\bf x}_i, y_i)\}_{i=1}^n$, which could be any prediction model~\cite{angelopoulos2021uncertainty, barber2021predictive}, CP relies on a negatively-oriented conformity score $s_n({\bf x},y): {\cal X}\times {\cal Y}\rightarrow \mathbb{R}$, which measures how well the prediction produced by the fitted model based on ${\cal D}_t$ conforms with the true value $y$~\cite{Papadopoulos2008Normalized}. A larger score indicates significant disagreement between the prediction and the true label $y$. 
By inverting the score function, the conformal prediction set is obtained as:
\begin{align}
    \mathcal{C}_{n}({\mathbf{x}})= \{y\in {\cal Y}: s_{n}(\mathbf{x},y)\leq q_{n}\} \label{eq:CP_set}
\end{align} 
where $q_{n}$ is an estimated $(1-\alpha)$-quantile of the score distribution. In standard CP, $q_{t}$ is set as the $\lceil (1-\alpha)(n+1) \rceil$-th smallest value of $\{s_n(\mathbf{x}_i,y_i)\}_{i=1}^n$~\cite{vovk2005algorithmic}. Under the exchangeability assumption of data in ${\cal D}_n$, the prediction set~\eqref{eq:CP_set} enjoys the finite-sample coverage guarantee:
$\mathbb{P}(y \in \mathcal{C}_n (\mathbf{x})) \geq 1-\alpha$.

Despite the appealing coverage guarantee, the exchangeability assumption is often violated in practice, especially in online settings where data arrives sequentially. In the graph setting, this challenge is amplified since node interdependencies naturally violate the i.i.d. assumption. To adapt CP to graphs, attempts have been made toward enforcing rather strict setting to accommodate data exchangeability; see, e.g., ~\cite{zargarbashi2023conformal, huang2023uncertainty, clarkson2023distribution}. However, these assumptions cannot be satisfied in the current streaming setting with arbitrary distributional shifts. To maintain valid coverage guarantee for the aforementioned graph-adaptive GP model, we devise adaptive mechanisms to adjust $q_t$ online in the following section, enabling robust coverage maintenance even when exchangeability is violated.

\section{Scalable Graph-adaptive GP ensembles with online CP }
\label{sec:pagestyle}
\subsection{Scalable Graph GPs with Random Features (RFs)}\label{sec:ensemble-RF}
Vanilla GP-based prediction~\eqref{eq:y_pre} incurs cubic complexity in the number of observed nodal labels, which becomes unaffordable as the number of data samples grow in the online setting.
To effect scalability, we will rely on the RF approximation~\cite{rahimi2007random} to yield a novel graph-aware GP approximant. For any shift-invariant  $\kappa({\bf x})$, the Bochner's theorem yields\\
${\kappa}(\mathbf{x}-\mathbf{x}') = \int \pi_{{\kappa}} (\mathbf{v}) e^{j\mathbf{v}^\top (\mathbf{x} - \mathbf{x}')} d\mathbf{v} = \mathbb{E}_{\pi_{{\kappa}}} \left[e^{j\mathbf{v}^\top (\mathbf{x} - \mathbf{x}')} \right]$,
where $\pi_{{\kappa}}$, the Fourier transform of $\kappa$, is the normalized power spectral density which can be viewed as a pdf. Drawing i.i.d. samples $\{\mathbf{v}_i\}_{i=1}^D$ from $\pi_{{\kappa}}(\mathbf{v})$, the RF vector $\bbphi_{\mathbf{v}} (\mathbf{x}) :=\! \frac{1}{\sqrt{D}}\!\left[\sin(\!\mathbf{v}_1^\top\! \mathbf{x}), \cos(\!\mathbf{v}_1^\top\! \mathbf{x}), \ldots, \sin(\!\mathbf{v}_{\! D}^\top\! \mathbf{x}), \cos(\!\mathbf{v}_{\! D}^\top \!\mathbf{x})\right]^{\top}$
yields approximation $\bar{\kappa}(\mathbf{x}, \mathbf{x}') \approx \bbphi_{\mathbf{v}}^{\top}(\mathbf{x})\bbphi_{\mathbf{v}}(\mathbf{x}')$. Then, one can obtain the RF-based linear function approximant for $f(\mathbf{x})$ as: $\check{f}(\mathbf{x}) =  \bbphi_{\mathbf{v}}^{\top}(\mathbf{x})\bbtheta, \bbtheta\sim {\cal N}({\bf 0}_{2D}, \sigma_\theta^2{\bf I}_{2D})$

Accounting for the graph topology via~\eqref{eq:h_N}, the latent $h$ conforms to the following generative model per node $n$\vspace{-0.1cm}
\begin{align}
\check{h}_n = \tilde{\bbphi}_{n}^\top\bbtheta, \quad \bbtheta \sim \mathcal{N}(\mathbf{0}_{2D}, \sigma_\theta^2\mathbf{I}_{2D})
\end{align}
 where $\tilde{\bbphi}_{n}$ is the $n$th column of $\tilde{\bbPhi}_N := \bbPhi_N {\bf P}_N^\top $ with $\bbPhi_N:=[\bbphi_{\mathbf{v}}(\mathbf{x}_1), \ldots, \bbphi_{\mathbf{v}}(\mathbf{x}_N)]$. Note that $\tilde{\bbPhi}_N$ are the average of the RF-based nodal features over the 1-hop neighbors of node $n$, in line with the discussion in~\cite{ng2018bayesian}.

With the Gaussian likelihood, one can obtain the posterior $p(\bbtheta|{\bf y}_n, {\cal G}) = {\cal N}(\bbtheta; \hat{\bbtheta}_n, \bbSig_n)$,  which can be used to predict for $y_{n+1}$ and is amenable to recursive Bayesian update at the complexity of $\mathcal{O}(D^2)$ per iteration when the true value of $y_{n+1}$ is revealed~\cite{lu2020ensemble,polyzos2021ensemble}.



\vspace{0.2cm}
\noindent {\bf Ensembling (E) GPs for adaptivity.}
To enhance adaptivity and robustness, we employ an ensemble of $M$ GP kernels, each maintaining its own parameter vector $\bbtheta^{(m)}$ with posterior
${p}(\bbtheta^{(m)}| \mathcal{G}, {\bf y}_n) = \mathcal{N}(\bbtheta^{(m)}; \hat{\bbtheta}_{n}^{(m)}, \bbSig_{n}^{(m)})$. Each model is associated with a weight $w_n^{(m)}: = \mathbb{P}(m|\mathcal{G}, {\bf y}_n)$ to assess its contribution adapted to the data on the fly.

To predict for the label at node $n+1$, each model $m$ provides prediction $p(y_{n+1}|\mathcal{G}, {\bf y}_n, m) = \mathcal{N}(y_{n+1}; \hat{y}_{n+1|n}^{(m)}, \sigma_{n+1|n}^{2,(m)})$,
where $\hat{y}_{n+1|n}^{(m)} \!\!= \tilde{\bbphi}_{n+1}^{(m)\top}\!\!\hat{\bbtheta}_{n}^{(m)}$ and $\sigma_{n+1|n}^{2,(m)} = \tilde{\bbphi}_{n+1}^{(m)\top}\bbSig_{n}^{(m)}\tilde{\bbphi}_{n+1}^{(m)}$.
Based on the sum-product probability rule, the EGP-based ensemble prediction is given by the Gaussian mixture (GM)\vspace{-0.3cm}
\begin{align}
p(y_{n+1}|\mathcal{G}, {\bf y}_n)\! =\!\! \sum_{m=1}^M w_n^{(m)}\mathcal{N}(y_{n+1}; \hat{y}_{n+1|n}^{(m)} , \sigma_{n+1|n}^{2,(m)}) \;. \label{eq:GM}
\end{align}\vspace{-0.5cm}\\
\noindent
There are various ways to form the prediction set for $y_{n+1}$ based on such a GM pdf. For the ease of computation, our approach is to approximate it using a single Gaussian pdf $\check{p}(y_{n+1}|\mathcal{G}, {\bf y}_n) =\mathcal{N}(y_{n+1}; \bar{y}_{n+1|n} , \bar{\sigma}_{n+1|n}^{2}) $, based on which one can readily obtain the BCS as in~\eqref{eq:BCS}. To obtain the moments $\{\bar{y}_{n+1|n} , \bar{\sigma}_{n+1|n}^{2}\}$, one can minimize the Kullback–Leibler (KL) divergence between the approximated Gaussian and the GM~\eqref{eq:GM}, which boils down to moment matching, yielding
\begin{align}
\bar{y}_{n+1|n}  &= \sum_{m=1}^M w_n^{(m)} \hat{y}_{n+1|n}^{(m)},\label{eq:ensemble-mean} \\
\bar{\sigma}_{n+1|n}^{2} &= \sum_{m=1}^M w_n^{(m)} \left[\sigma_{n+1|n}^{2,(m)} + (\hat{y}_{n+1|n}^{(m)} - \bar{y}_{n+1|n})^2\right]\label{eq:ensemble-var}
\end{align}
Upon observing $y_{n+1}$, one can use Bayes' rule to update 
\begin{align}
&w_{n+1}^{(m)} \propto w_n^{(m)}p(y_{n+1}|\mathcal{G}, {\bf y}_n, m) \\
& {p}(\bbtheta^{(m)}| \mathcal{G}, {\bf y}_{n+1}) \propto {p}(\bbtheta^{(m)}| \mathcal{G}, {\bf y}_{n})p(y_{n+1}|\bbtheta^{(m)})
\end{align}
which can be implemented efficiently at the complexity of $\mathcal{O}(M \cdot D^2)$  per iteration~\cite{lu2020ensemble,polyzos2021ensemble}.

\subsection{Online CP with RF-based EGPs over graphs}

While the RF-based EGPs can construct BCSs via Eq.~\eqref{eq:BCS}, these intervals may fail to maintain the target coverage $\beta$ when model assumptions are violated or data distributions shift over time. To ensure provably valid coverage despite model misspecification and non-exchangeability, we integrate our EGP framework with online conformal prediction (OCP), which adaptively adjusts prediction thresholds based on empirical coverage rates.

Leveraging the Bayesian nature of our EGP predictor, we will employ the negative predictive log-likelihood (NPLL) as the nonconformity score~\cite{xu2025online}. For the ease of obtaining the prediction sets, we will evaluate the NPLL using the approximated Gaussian $\check{p}(y_{n+1}|\mathcal{G}, {\bf y}_n)$, yielding 
\begin{align}
s_{n+1}(y) = \frac{1}{2}\log(2\pi\bar{\sigma}_{n+1|n}^{2}) + \frac{(y - \bar{y}_{n+1|n})^2}{2\bar{\sigma}_{n+1|n}^{2}} \label{eq:ensemble-score}
\end{align}
where $\bar{y}_{n+1|n}$ and $\bar{\sigma}_{n+1|n}^{2}$ are the ensemble predictions from~\eqref{eq:ensemble-mean}-\eqref{eq:ensemble-var}. The CP set at node $n+1$ is $\mathcal{C}_{n+1} = \{y: s_{n+1}(y) \leq q_{n}\}$, with threshold $q_n$ adaptively updated via: $q_{n+1} = q_n - \eta(\alpha - \mathbb{I}(s_{n+1}(y_{n+1}) > q_n))$, where $\eta > 0$ is the learning rate and $\mathbb{I}(\cdot)$ indicates miscoverage.

\begin{table*}[h]
\centering
\caption{Coverage Performance (\%) Across Datasets and Model Types}
\label{tab:all_models_comparison}
\begin{tabular}{l|cc|cc|cc|cc}
\hline
\multirow{2}{*}{Method} & \multicolumn{2}{c|}{Heteroscedastic} & \multicolumn{2}{c|}{Linear} & \multicolumn{2}{c|}{California Housing} & \multicolumn{2}{c}{Bike Sharing} \\
& Coverage & Width & Coverage & Width & Coverage & Width & Coverage & Width \\
\hline
EGP-CP & $88.08 \pm 1.44$ & $0.97$ & $88.54 \pm 1.17$ & $0.89$ & $90.81 \pm 0.33$ & $1.61$ & $90.40 \pm 0.98$ & $2.21$ \\
RBF-CP & $88.32 \pm 1.45$ & $0.91$ & $88.78 \pm 1.25$ & $0.91$ & $90.75 \pm 0.53$ & $1.64$ & $90.48 \pm 0.93$ & $2.22$ \\
\hline
EGP-SNAPS & $83.99 \pm 1.63$ & $0.87$ & $84.72 \pm 1.34$ & $0.78$ & $84.69 \pm 0.85$ & $1.58$ & $83.49 \pm 1.15$ & $2.08$ \\
RBF-SNAPS & $84.30 \pm 1.65$ & $0.81$ & $85.05 \pm 1.44$ & $0.81$ & $84.84 \pm 0.87$ & $1.61$ & $83.32 \pm 1.61$ & $2.08$ \\
\hline
EGP-OCP & $\mathbf{89.61 \pm 0.51}$ & $0.98$ & $\mathbf{89.66 \pm 0.33}$ & $0.94$ & $\mathbf{90.62 \pm 0.31}$ & $1.68$ & $\mathbf{90.32 \pm 0.79}$ & $2.23$ \\
RBF-OCP & $89.33 \pm 0.55$ & $0.97$ & $89.50 \pm 0.39$ & $0.96$ & $90.63 \pm 0.48$ & $1.71$ & $90.44 \pm 0.82$ & $2.24$ \\
\hline
EGP-BCS & $91.80 \pm 1.08$ & $1.10$ & $93.63 \pm 0.54$ & $1.11$ & $71.90 \pm 1.82$ & $1.26$ & $63.70 \pm 0.84$ & $1.16$ \\
RBF-BCS & $90.69 \pm 1.29$ & $1.08$ & $92.58 \pm 0.57$ & $1.08$ & $70.67 \pm 1.90$ & $1.26$ & $62.22 \pm 0.66$ & $1.16$ \\
\hline
\end{tabular}
\end{table*}

\noindent {\bf Coverage Guarantee.}
For graph  ${\cal G}$ with $N$ nodes where labels arrive sequentially, with constant learning rate $\eta > 0$, if the nonconformity score $s_n(y) \in [0, B]$ and initial threshold $q_0 \in [0, B]$, the long-run coverage of our online CP satisfies:
\begin{align}
\left|\frac{1}{N}\sum_{n=1}^{N} \mathbb{I}(y_n \in \mathcal{C}_{n}) - (1-\alpha)\right| \leq \frac{B + \eta}{\eta N} \;.
\end{align}
This bound, derived based on Theorem 1 in~\cite{angelopoulos2024online},  shows that the coverage converges to the target level $(1-\alpha)$ at rate $O(1/N)$, ensuring asymptotic validity of the OCP procedure.

\section{NUMERICAL Results}
We evaluate the proposed method on two synthetic and two real-world datasets for graph-based inference. The first synthetic dataset has heteroscedastic noise whose variance increases with feature magnitude, while the second synthetic dataset features primarily linear relationships with small homoscedastic noise. The two real datasets are given by the \texttt{California Housing} and \texttt{Bike Sharing} datasets. For all the datasets,
we construct $k$-NN graphs ($k=6$) from feature similarity using Euclidean distance, which is a common practise in the literature.  We partition the data into initial training set $\mathcal{D}_{\text{init}} := \{(\mathbf{x}_n, y_n)\}_{n=1}^{n_{\text{init}}}$ and test stream $\mathcal{D}_{\text{stream}} = \{(\mathbf{x}_n, y_n)\}_{n=n_{\text{init}}+1}^{N}$. All datasets use 30\% initial training and 70\% streaming test splits with normalized features. Unlike traditional CP that requires a separate calibration set, we employ a \textit{pure online setting} where the initial training set serves dual purposes: (i) training the ensemble GP models following Section~\ref{sec:ensemble-RF}, and (ii) computing initial nonconformity scores to set the conformal threshold $q_0$. The conformal threshold is initialized as $q_0 = \text{Quantile}_{1-\alpha}(\{s_n( y)\}_{n=1}^{n_{\text{init}}})$, where $s_n(y)$ are the nonconformity scores on the initial training data.

We evaluate our proposed OCP with graph-aware EGP model consisting of three kernels (RBF, Mat\'ern-2.5, and Mat\'ern-1.5) using $D = 400$ RFs and adaptive threshold $\eta=0.01$. To validate the merits of using ensemble models, we compare against single GP-based prediction model with the best-performing RBF kernel. Further, OCP is compared with three baselines, namely, traditional CP (fixed threshold), SNAPS (graph-aware fixed threshold)~\cite{song2024similarity}, and vanilla GP-based BCS~\eqref{eq:BCS}. All experiments target 90\% coverage ($\alpha=0.1$) over 50 independent runs. Performance is measured by: (1) empirical coverage rate (fraction of test points containing true labels; and (2) average interval width ( narrower width indicates more efficient UQ). Optimal performance achieves near-90\% coverage with tight prediction intervals.

\noindent{\bf Results Analysis.}
EGP-OCP achieves the most reliable coverage performance, consistently approaching the 90\% target across all datasets while maintaining exceptional stability (std $< 1\%$). The adaptive threshold mechanism proves crucial for handling distribution shifts, as traditional CP exhibits higher variance despite comparable mean coverage. EGP model provide consistent robustness gains over single GP-based counterpart across all set predictors, with the stability improvements being most pronounced for OCP, confirming that kernel diversity enhances prediction reliability in streaming scenarios.
SNAPS undercovers across all datasets (83--85\%), as its exchangeability assumption is violated both spatially (graph dependencies) and temporally (distribution shifts), failing to capture the non-exchangeable nature of streaming graph data. On the other hand, BCS performs reasonably on synthetic data but fails dramatically on real-world datasets (62--72\% coverage), validating the effect of model mis-specification on the coverage performance. The coverage-width trade-off analysis shows that OCP achieves superior statistical guarantees with minimal efficiency cost, requiring only marginally wider intervals than under-performing baselines. These results establish EGP-OCP as the optimal approach for online UQ over graphs, combining near-target coverage with robust cross-domain performance.

\section{CONCLUSIONS}
\label{sec:majhead}

This work put forth a novel scalable graph-aware GP variant by incorporating the topology information into the RF approximation. To effect adaptivity to online setting where nodal labels arriving on the fly, an ensemble of graph-adaptive GPs are employed, where the per-GP weight and parameter posterior are amenable to recursive Bayesian update with scalability. To further combat against model mis-specification and data distributional shift, the resulting graph-based EGP based set predictor is wed with the OCP framework, where the key threshold parameter is adaptively adjusted based on coverage feedback. Experimental results demonstrated that the graph-adaptive EGP-OCP achieves superior coverage relative to existing baselines across both synthetic and real graph-based datasets, validating its effectiveness for diverse streaming scenarios while maintaining computational efficiency suitable for real-time deployment.


\printbibliography

\end{document}